\documentclass[preprint,authoryear,11pt,a4paper]{elsarticle}
\usepackage[top=0.75in, bottom=0.75in, left=0.75in, right=0.75in]{geometry}

\usepackage{amsmath,amsthm}
\usepackage{amssymb}
\usepackage{amsfonts}
\usepackage{graphicx}
\graphicspath{{graphics/}}
\usepackage{fnpct} 

\usepackage{xcolor}
\definecolor{darkblue}{rgb}{0.0,0.5,0.5}
\definecolor{blue}{rgb}{0.0,0.5,0.68}
\usepackage[colorlinks]{hyperref}
\hypersetup{colorlinks,breaklinks,linkcolor=darkblue,urlcolor=darkblue,anchorcolor=darkblue,citecolor=darkblue}
\usepackage{booktabs}
\usepackage{threeparttable}
\usepackage{multirow}
\usepackage{hhline}
\usepackage{lscape}
\usepackage{subcaption}
\usepackage{epstopdf}

\usepackage{microtype}
\usepackage{lineno}

\journal{}

\begin{document}

\begin{frontmatter}

\title{Joint predictions of multi-modal ride-hailing demands: a deep multi-task multi-graph learning-based approach}
\author[1]{Jintao Ke}
\author[1]{Siyuan Feng}
\author[1]{Zheng Zhu\corref{cor1}}
\ead{zhuzheng@ust.hk}
\cortext[cor1]{Corresponding author}
\author[1]{Hai Yang}
\author[2,3]{Jieping Ye}

\address[1]{Department of Civil and Environmental Engineering, the Hong Kong University of Science and Technology, Kowloon, Hong Kong}
\address[2]{Department of Computational Medicine and Bioinformatics, University of Michigan, Ann Arbor, United States}
\address[3]{AI Labs, Didi Chuxing, Beijing, China}



\begin{abstract}

Ride-hailing platforms generally provide various service options to customers, such as solo ride services, shared ride services, etc. It is generally expected that demands for different service modes are correlated, and the prediction of demand for one service mode can benefit from historical observations of demands for other service modes. Moreover, an accurate joint prediction of demands for multiple service modes can help the platforms better allocate and dispatch vehicle resources. Although there is a large stream of literature on ride-hailing demand predictions for one specific service mode, little efforts have been paid towards joint predictions of ride-hailing demands for multiple service modes. To address this issue, we propose a deep multi-task multi-graph learning approach, which combines two components: (1) multiple multi-graph convolutional (MGC) networks for predicting demands for different service modes, and (2) multi-task learning modules that enable knowledge sharing across multiple MGC networks. More specifically, two multi-task learning structures are established. The first one is the regularized cross-task learning, which builds cross-task connections among the inputs and outputs of multiple MGC networks. The second one is the multi-linear relationship learning, which imposes a prior tensor normal distribution on the weights of various MGC networks. Although there are no concrete bridges between different MGC networks, the weights of these networks are constrained by each other and subject to a common prior distribution. Evaluated with the for-hire-vehicle datasets in Manhattan, we show that our propose approach outperforms the benchmark algorithms in prediction accuracy for different ride-hailing modes.  
\end{abstract}


\end{frontmatter}

\section{Introduction}
Ride-hailing services, offering customers with door-to-door ride services at any time and anywhere, have experienced explosive growth in recent years. One advantage of ride-hailing companies over traditional street-hailing taxi companies is that ride-sourcing companies can track and record the real-time trip information from both passenger side and driver side. Based on this information, the platform can discover the representative demand-supply patterns and predict passenger demand over time and over space (number of ride requests originating from one specific zone during one time interval). An accurate short-term prediction of passenger demand serves as a foundation of many operating strategies that aim to improve system efficiencies, such as surge pricing, vehicle dispatching, and vacant vehicle re-allocation, etc.

Most of the existing studies focus on predicting region-level ride-hailing passenger demand for one service mode \citep{yao2018deep, yao2019revisiting, ke2017short, geng2019multi, geng2019spatiotemporal}. They partition the examined city into various regular regions (squares or hexagons) or irregular regions (on the basis of administrative or geographical properties) and predict the near-future passenger demand in each region. However, in actual operations, ride-hailing companies commonly provide diversified ride services to customers with different interests. For example, solo ride services (such as UberX, Lyft, Didi Express), which dispatch one vehicle to serve one passenger at each time, are preferable by customers who are more inclined to time or feel uncomfortable about sharing rides with others. By contrast, shared ride services (such as UberPool, Lyft Shared, Didi ExpressPool), which allow one driver to pick-up and drop-off two or more passengers in each ride with a discounted trip fare, are provided to customers who are more inclined to money. Some platforms even provide luxury ride services, such as Uber Black, to customers who are willing to pay for a better car environment. Moreover, many passengers do not stick to one service mode; instead, they may switch among different service modes in different circumstances \citep{lavieri2019investigating}. For example, during peak hours, due to supply limitations, the platforms may implement surge pricing and raise the trip fare, such that passengers are more prone to use shared rides with a relatively low trip fare. This indicates that demands for different service modes interact with each other, thereby the historical observations of demands for one service mode can provide valuable information to the prediction of demands for other service modes. 

Meanwhile, the platforms also have a strong desire for an accurate joint prediction of demands for multiple service modes, which help them better allocate and dispatch vehicle resources. For example, when the platform predicts that the demand for solo ride services will be substantially greater than the supply of regular vehicles in one region and there are sufficient idle luxury vehicles nearby, it can mitigate passenger queuing by dispatching luxury vehicles to serve solo ride passengers with free service upgrades. However, although the spatial-temporal prediction for ride-hailing demand has been examined for many years, most of the previous studies focused on prediction for one specific service mode. It remains unsolved and challenging in how to provide an accurate joint prediction of multi-modal ride-hailing demands, by a unified approach that can simultaneously model the spatial-temporal dependencies and knowledge sharing across prediction tasks. 

To tackle this challenge, this study proposes a novel multi-task multi-graph learning approach. The approach views the prediction of each ride-hailing mode demand as one task. For each task, we propose a multi-graph convolutional (MGC) network to capture the non-Euclidean spatial-temporal dependencies among different regions based on both geographical and semantic aspects. Multiple graphs are developed, including a distance graph that models the pair-wise distance between each two regions, a neighborhood graph that indicates whether two regions are adjacent to each other, a functionality graph that characterizes the functional similarity between each two regions, and a mobility pattern graph that describes the correlation of the historical demand trends between each two regions. 

On the basis of various MGC networks, we design two multi-task learning structures to share knowledge across different spatial-temporal prediction tasks. The first one is the regularized cross-task (RCT) learning, which builds concrete crossed connections between the inputs and outputs of different tasks, such that prediction of one service mode demand can take advantage of information from other service modes. In the objective function, to avoid over-fitting issues due to model complexity, we penalize the inter-task weights and intra-task weights with different intensities. The second structure is the multi-linear relationship (MLR) learning. Instead of using inter-task weights to concretely link different tasks, MLR assumes that the intra-weights of various MGC networks are subject to a common tensor normal prior distribution. Therefore, weights in different networks are restrained by each other, and different tasks learn to share knowledge. Based on multi-modal demand prediction experiments with actual ride-hailing data in Manhattan, New York, the proposed framework outperforms the benchmark algorithms. In summary, this paper makes the following contributions: 

\begin{itemize}
\item We propose a novel multi-task multi-graph learning approach to enable the joint prediction of multi-modal ride-hailing demands as well as other spatial-temporal joint prediction tasks. 
\item Two multi-task learning methods, namely RCT learning and MLR learning, are proposed to share knowledge across the MGC networks for different prediction tasks. 
\item We conduct extensive experiments on the actual ride-hailing dataset in Manhattan which contains both solo and shared ride services. We show that the proposed approach outperforms the state-of-art algorithms, and the use of multi-task learning structures can improve predictive accuracy in different spatial-temporal prediction tasks. 
\end{itemize}

\section{Literature review}
The forecasting of ride-hailing demands belongs to the huge family of spatial-temporal predictions. In this section, we provide a thorough review of conventional and advanced approaches for spatial-temporal prediction of travel demand as well as other traffic states (such as flow, speed and density). Of particular focus is the emerging multi-task learning-based approaches that enable us to predict multi-modal ride-hailing demands or other traffic-related measurements simultaneously.

\subsection{Conventional spatial-temporal approaches}
The prediction of short-term transportation measurements was brought to the academic field in 1979 when the autoregressive integrated moving average (ARIMA) model was introduced to predict traffic flows \citep{ahmed1979analysis}. The time series ARIMA approach has been refined over time \citep{levin1980forecasting, hamed1995short, billings2006application}. Other statistical models and machine learning models were also proposed to solve prediction problems of traffic flow, traffic incidents, and travel demand. Conventional prediction approaches include regressions \citep{kamarianakis2010characterizing, battifarano2019predicting}, Kalman filtering models \citep{okutani1984dynamic, lu2014short}, Bayesian network (BN) models \citep{zhu2016short}, Neural network models \citep{park1998forecasting, zheng2006short}, K-nearest neighbor algorithm \citep{tak2014real} and so on.

The majority of these approaches treat the predicted transportation states as univariate time series, ignoring the nature of spatial correlations in transportation systems. Some researchers have considered spatial-temporal covariates into traditional approaches for traffic states and travel demand predictions. \cite{yin2002urban} considered upstream time series traffic flows to predict downstream traffic states via a fuzzy-neural model. \cite{sun2006bayesian} adopted a Gaussian BN model to predict near-future traffic flow with both local and upstream volumes. \cite{zhu2019conditional} incorporated the joint probability distributions of traffic flows at nearby sensor stations into traffic speed prediction. Spatial-temporal covariates were also utilized via conventional approaches for the predictions of travel time \citep{wu2004travel}, rail demand \citep{jiang2014short}, metro demand \citep{ni2016forecasting}, etc. Although conventional approaches have alleviated the difficulties in forecasting the stochasticity of transportation states, a common limitation is that only the nearby spatial information was included in these models. With traditional model structures and estimation algorithms, it can be difficult to incorporate useful distant information into predictions.

\subsection{Deep learning spatial-temporal approaches}
In recent years, deep learning-based approaches have been widely used in transportation state predictions. Designed for research tasks such as image recognition, convolutional neural networks (CNNs) are capable of capturing high-order spatial-temporal correlations in transportation prediction problems. Spatial-temporal transportation states are naturally regarded as a series of images by dividing the study area into small regions or zones. And following this approach, researchers have utilized CNNs in various prediction tasks, including speed evaluation \citep{ma2015large}, bike usage prediction \citep{zhang2016dnn}, ride-hailing demand-supply prediction \citep{ke2018hexagon} and so on. Recurrent neural networks (RNNs) and their extensions such as long short-term memory (LSTM) are well fit for processing time series data streams. \cite{xu2017real} applied LSTM
to predict taxi demand in New York city. Some researchers integrated RNNs with CNNs to make full use of spatial-temporal information to forecast short-term ride-hailing demand \citep{ke2017short}, traffic flow \citep{wu2016short, yu2017spatiotemporal} and bike flow \citep{zhang2018predicting}.

Based on but not limited to the mechanism of CNNs and RNNs, there have been extensions on the integrated deep learning algorithms. \cite{liu2019contextualized} developed a contextualized spatial-temporal network, which captures a local spatial context, a temporal evolution context, and a global correlation context, to predict taxi demand. \cite{geng2019spatiotemporal} proposed a spatial-temporal MCG (ST-MCG) model that utilizes non-Euclidean correlations for ride-hailing demand prediction. Based on an encoding-decoding structure between CNNs and ConvLSTMs, \cite{zhou2018predicting} developed an attention-based deep neural network to forecast multi-step passenger demand for bike and taxi.

\subsection{Multi-task learning-based approaches}
The aforementioned conventional and advanced approaches greatly enhance the capability of urban-wise mobility prediction and evaluation. The superiority in prediction accuracy with a specific transportation state (e.g. traffic flow and travel demand) forecasting task has been demonstrated in previous studies. Since transportation states can be correlated with each other, researchers become interested in the simultaneous prediction of multiple states. For instance, joint-prediction of morning and evening commute demands may be more accurate than single demand predictions due to the positive correlation between the two types of commute demands.

In machine learning approaches, multi-task learning is a good solution to joint prediction problems. Multi-task learning is a paradigm that aims to leverage useful information contained in multiple learning tasks for improving the performance of various tasks \citep{zhang2017survey}. A deep multi-task learning model attempts to learn the correlated representation in the feature layers and independent classifiers in the classifier layer without affecting the relationships of the tasks \citep{long2017learning}. Nowadays, substantial research efforts are dedicated to the application of multi-task deep learning algorithms for the simultaneous prediction of correlated transportation states. \cite{kuang2019predicting} embedded the common features of taxi pickup demand and taxi dropoff demand via an attention-based LSTM model, and jointly predicted the two taxi demands via a 3D residual deep neural network. \cite{geng2019multi} proposed a modality interaction mechanism to learn the interactions among different region-wise graph representations in MGCs. \cite{zhang2019short} proposed a multi-task temporal CNN approach for zone-level travel demand prediction.

However, little efforts have been directed towards the joint prediction of demands for multiple service modes in ride-hailing systems. Concerning the correlations among different ride-hailing service modes, it is meaningful to explore suitable ways to share knowledge across the prediction tasks for various demands.

\section{Preliminaries}
In this section, we first give explicit definitions to several key concepts and then formulate the multi-modal ride-hailing demands prediction problem.  

\subsection{Region partition}

It is a common way in the literature to partition the examined area into various regular rectangles. This allows easy implementations of stylized spatial-temporal prediction models, such as CNNs, RNNs, and combinations of CNNs and RNNs, etc. There are also some studies \citep[e.g.][]{ke2018hexagon} dividing the examined city into various regular hexagonal grids due to the fact that hexagons have an unambiguous neighborhood definition, a smaller edge-to-area ratio (smoother boundaries) and nice isotropic properties. However, in actual city management, regulators and planners are more prone to divide the city into various irregular grids, according to their administrative and geographical properties. In this paper, we use Manhattan, New York City as the testbed, and partition it into 63 zones according to the administrative zone system of New York Taxi and Limousine Commission (TLC). As for the temporal dimension, each day is uniformly divided into intervals with equal length time slices (e.g. one hour). 

On the basis of the administrative region partitions, we build a weighted graph with nodes referring to the zones and edges characterizing the inter-zone relationships; thereby, zones are fully connected with each other in this graph (i.e. any two nodes have a connection via a link).. Let $G(V, E, \boldsymbol{A})$ denote the weighted graph, where $V$ is the set of zones, $E$ is the set of edges, and $\boldsymbol{A} \in \mathbb{R}^{|V| \times |V|}$ is the adjacent matrix with each element indicating the relationship between two zones. 

\subsection{Research problem}

In this paper, we target at predicting multi-modal region-level ride-hailing passenger demands in a short time interval. Suppose the platform provides a total of $M$ ride-hailing service modes (such as expresses, luxury, shared ride service, etc.). Let $x_{i, m}^{t}$ denote the number of passenger requests (passenger demand) for service mode $m$ in zone $i$ during time interval $t$, and $\boldsymbol{X}_{m}^{t}$ denote passenger demands for service mode $m$ in all zones at time interval $t$.  As examined in many previous studies \citep[e.g.][]{ke2017short, geng2019spatiotemporal,yao2019revisiting} the problem of region-level ride-hailing demand prediction for one service mode $m$ can be formulated as a single-task problem as follows, 

\textit{Definition 1}.  (ride-hailing demand prediction) Given the historical observations of ride-hailing demand for service mode $m$ before the current time interval $t$, that is $[\boldsymbol{X}_{m}^{t-T}, ..., \boldsymbol{X}_{m}^{t}]$, the problem is to predict the spatial-temporal ride-hailing demand for service mode $m$ in the next time interval, that is, $\boldsymbol{X}_{m}^{t+1}$. $T$ is the number of historical time intervals used for the prediction.  

As aforementioned, it is naturally expected that demand prediction for one mode can benefit from the historical observations of demands for other modes. With this knowledge in mind, we formulate a multi-task learning problem that simultaneously predicts ride-hailing demands for all service modes by taking advantage of the historical demands for all service modes. The problem is formally defined as,

\textit{Definition 2}. (multi-modal ride-hailing demands prediction) Given the historical observations of ride-hailing demands for service modes $[\boldsymbol{X}_{m}^{t-T}, ..., \boldsymbol{X}_{m}^{t}], \forall m \in \{1, ..., M\}$, the problem is to forecast the spatial-temporal ride-hailing demand for multiple service modes $\boldsymbol{X}_{m}^{t+1}, \forall m \in \{1, ..., M\}$.  

As pointed out by \cite{zhang2017survey}, one important issue in multi-task learning is how to share knowledge among various tasks. In what follows, we will present a multi-task multi-graph learning approach that spells out the concrete ways to share knowledge among different service modes for a better multi-modal demand prediction. 

\section{A deep multi-task multi-graph learning approach}

In our proposed approach, we first capture both geographical and semantical non-Euclidean relationships among zones in multiple graphs. It is worth to mention that the graphs for different service modes are not identical, since some graphs characterize the mobility patterns (trends of historical demand), which are different across service modes. For each service mode, we then implement an MGC network to predict its region-level (i.e. zone-level) demand on the basis of its corresponding graphs. Finally, we propose two multi-task learning structures, the RCT learning and MLR learning, that specify the ways to share knowledge across different tasks (namely, predictions for different service modes). 

\subsection{Spatial dependence and multi-graphs}
In an MGC network, geographical and semantic relationships among zones are represented by the graph structure and its associated adjacent matrices. Now we construct three common graphs that are shared by all service modes (the neighborhood graph $G_N(V, E, \boldsymbol{A}_N)$, distance graph $G_D(V, E, \boldsymbol{A}_D)$, and functionality graph $G_F(V, E, \boldsymbol{A}_F)$), and one specific graph that is diverse across different service modes, i.e. the mobility pattern graph $G_P^m(V, E, \boldsymbol{A}_P^m)$. Formally, $\boldsymbol{A}_N$ and $\boldsymbol{A}_D$ are given by, 

\begin{equation}
[\boldsymbol{A}_N]_{i,j} = \begin{cases}
               1\text{, if zone i and j are adjacent}\\
               0\text{, otherwise}
            \end{cases}
\end{equation}

\begin{equation}
[\boldsymbol{A}_D]_{i,j} = \frac{1}{Dist(lng_i, lat_i, lng_j, lat_j)} 
\end{equation}
where $lng_i$, $lat_i$ are the longitude and latitude of the central point of zone $i$, $Dist(\cdot)$
calculates the straight-line distance between point $(lng_i, lat_i)$ and $(lng_j, lat_j)$, $[\boldsymbol{A}_N]_{i,j}$ refers to the element of adjacent matrix $A_N$ in the $i$th row and $j$th column. Clearly, the shorter the straight-line distance between the centers of two zones, the larger the weight associated with these two zones in the distance graph (the stronger the relationship). These two graphs can well capture the pair-wise geographical relationships between zones.  

In addition to having geographical relationships, different zones may be correlated with each other in a semantic manner. Usually, zones in a city have different functionalities or land-use properties: some are business zones, while others are residential zones. The ride-hailing demands in two zones with similar functionalities can be strongly correlated, even though they are far away from each other geographically. With this knowledge in mind, we formulate the functionality  graph by, 

\begin{equation}
[\boldsymbol{A}_F]_{i,j} = \frac{1}{\sqrt{(\boldsymbol{s}_i - \boldsymbol{s}_j) (\boldsymbol{s}_i - \boldsymbol{s}_j)^T}}
\label{eq:fun}
\end{equation}

\noindent where $\boldsymbol{s}_i$, $\boldsymbol{s}_j$ are the vector of functionalities of zone $i$ and $j$. The vector of each zone \footnote{obtained from Smart Location Database: https://www.epa.gov/smartgrowth/smart-location-mapping } includes the number of households without private cars, the density of houses, the density of population, the density of employments, lengths of road network per square kilometers, and average distances to metro stations, etc. 

It is also generally expected that zones with similar mobility patterns (represented by historical demand trends) may share common characteristics and provide useful predictive information to each other \citep{yao2018deep}. Historical demand trends are different across service modes, and therefore we establish mode-specific mobility pattern graphs. For a specific service mode $m$, we have, 

\begin{equation}
[\boldsymbol{A}_P^m]_{i,j} = \frac{\text{Cov}(q_i^m, q_j^m)}{\sqrt{\text{Var}(q_i^m) \text{Var}(q_j^m)} } 
\label{eq:cor}
\end{equation}

\noindent where $q_i^m$, $q_j^m$ are the long-term historical trends (vectors) of ride-hailing demand for service mode $m$ in zone $i$ and $j$, respectively, Cor($\cdot$, $\cdot$) calculates the correlation of two time series vectors, Var($\cdot$) calculates the variance of one time series vector. 
 
 \subsection{Multi-graph convolutions}
 
In the past few years, researchers have developed various types of graph neural networks. These networks can be roughly categorized into two groups: spectral graph convolutional networks that transform signals from graph domain to Fourier domain through a graph Laplacian, and spatial graph convolution networks that directly operate in the graph domain. In this paper, we mainly consider the spectral convolutions. To efficiently transform signals, \cite{defferrard2016convolutional} employed a Chebyshev polynomial to approximate the graph Laplacian, and \cite{kipf2016semi} further simplified the graph Laplacian by re-normalizing a first-order Chebyshev polynomial. The latter method has a neat mathematical form and is widely used in many applications, such as node classifications in scholar networks and link prediction in social networks. In the spirit of this work and on the basis of the aforementioned multi-graphs, we formulate an MGC in the prediction for service mode $m$ by, 

\begin{equation}
\begin{split}
&\mathcal{F}_W^m (\boldsymbol{X}; \boldsymbol{A}_N, \boldsymbol{A}_D, \boldsymbol{A}_F, \boldsymbol{A}_P^m) \\
&= \sigma \left( \sum_{r \in \{ N,D,F,P \}}  \widehat{\boldsymbol{A}_r^m } \boldsymbol{X} \boldsymbol{W}_{r,m} + b_m \right) 
\end{split}
\end{equation} 

\noindent where $\boldsymbol{W}_{r,m} \in \mathbb{R}^{f_i \times f_o}, \forall r \in \{ N,D,F,P \}$ are trainable weights, $\boldsymbol{X} \in \mathbb{R}^{|V| \times f_i}$ are input features, $f_i$ and  $f_o$ are the input and output feature dimensions, $\sigma(\cdot)$ is an activation function, $b_m$ is the intercept. Matrix $\widehat{\boldsymbol{A}_r^m }$ is determined before training and given by, 

\begin{equation}
\widehat{\boldsymbol{A}_r^m } = \left( \boldsymbol{D}_r^m \right) ^{-1/2} \widetilde{ \boldsymbol{A}_r^m}  \left( \boldsymbol{D}_r^m \right)^{-1/2}
\end{equation}

\noindent where $\widetilde{ \boldsymbol{A}_r^m} = \boldsymbol{A}_r^m + \boldsymbol{I}$ is the sum of adjacent matrix and an identity matrix to ensure that each node takes advantage of the historical observations of itself. $\boldsymbol{D}_r^m$ is the degree matrix, where $[\boldsymbol{D}_r^m]_{ij} = \sum_j [\widetilde{ \boldsymbol{A}_r^m}]_{ij}$. It can be shown that our MGC assigns different weights to multiple graphs, and uses the sum of the outputs of multiple graphs to generate the final output, in each service mode. Therefore, in one single graph convolution, we treat all trainable weights (for different graphs) as one weight matrix $\boldsymbol{W}_m = [...,\boldsymbol{W}_{r,m}, ...] \in \mathbb{R}^{\tilde{f_i } \times f_o}$, where $\tilde{f_i } = f_i * 4 $. 

\subsection{Regularized cross-task learning}

\begin{figure}[ht!]
\centering
\includegraphics[width=0.62\textwidth]{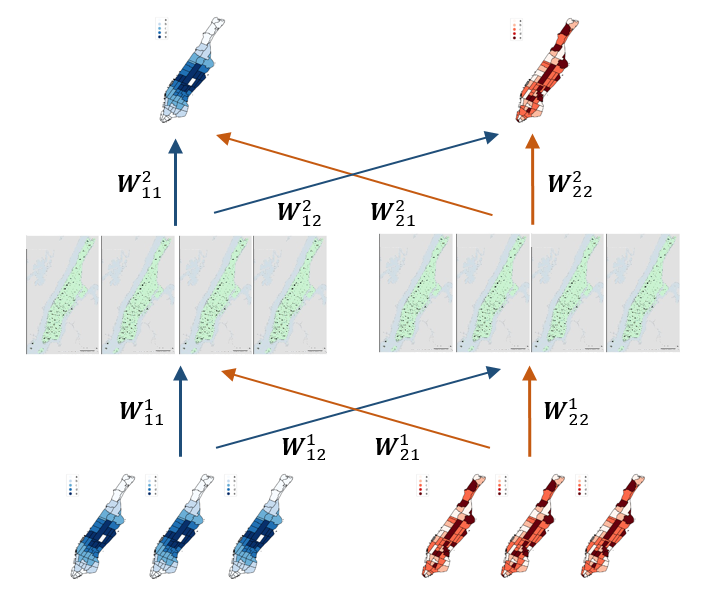}
\caption{Regularized cross-task learning}
\label{fig:cross-task}
\end{figure}

In this section, we propose a novel RCT learning structure that enables the predictions of different service modes to share knowledge with each other. To elaborate the key idea of RCT, we use Fig. \ref{fig:cross-task} as a demo, in which two basic three-layer networks are established to predict the ride-hailing demand for two service modes (mode 1 in blue color may represent solo service and mode 2 in red color may denote shared service). Let $\boldsymbol{W}^l_{m \rightarrow n}$ denote the trainable weight matrix (containing trainable weights for all graphs as mentioned above) that is associated with a graph convolution operation from service mode $m$ to service mode $n$ in the $l$th layer.  Without knowledge sharing (single-task learning), the network on the left directly maps the features of service mode 1 to its labels through two trainable weights $\boldsymbol{W}^1_{1 \rightarrow 1}$ and $\boldsymbol{W}^2_{1 \rightarrow 1}$; similarly, weights $\boldsymbol{W}^1_{2 \rightarrow 2}$ and $\boldsymbol{W}^2_{2 \rightarrow 2}$ are used to map the features of mode 2. This indicates that the networks for predicting different service modes are independent of each other. 

In RCT learning, we design a cross-task structure among networks for different service modes. Mathematically, the output of the network for service mode $m$ in layer $l$, denoted by $\boldsymbol{H}_m^{l+1}$ is given by, 

\begin{equation}
\boldsymbol{H}_m^{l+1} = \sum_{k \in \{ 1,...,M \}} \mathcal{F}_{\boldsymbol{W}^l_{k \rightarrow m}}^k (\boldsymbol{\boldsymbol{H}_k^l}; \boldsymbol{A}_N, \boldsymbol{A}_D, \boldsymbol{A}_F, \boldsymbol{A}_P^k)
\end{equation}

\noindent where convolution operation $\mathcal{F}_{\boldsymbol{W}^l_{k \rightarrow m}}^k$ maps from $\boldsymbol{H}_k^l$, namely, the inputs of the network for service mode $k$ in layer $l$, to $\boldsymbol{H}_m^{l+1}$, and is parameterized by $\boldsymbol{W}^l_{k \rightarrow m}$. We denote the weights that transform input to output within the same task as intra-task weights, and the weights that connect input and output of different tasks as inter-task weights. For example, in Fig. \ref{fig:cross-task}, $\boldsymbol{W}^1_{1 \rightarrow 1}$ and $\boldsymbol{W}^2_{1 \rightarrow 1}$ are intra-weights, while $\boldsymbol{W}^1_{1 \rightarrow 2}$ and $\boldsymbol{W}^2_{2 \rightarrow 1}$ are inter-weights. In this way, the prediction task for a service mode $m$ can take advantage of the information not only from its own features, but also from features of other service modes. However, RCT learning may greatly increase the number of weights, particularly when there are many service modes. To address this problem, we penalize the weights in the objective function by introducing the following regularization term: 

\begin{equation} \label{eq:j1}
J_1^l = \alpha  \sum_{i=1}^M \left\lVert  \boldsymbol{W}^l_{i \rightarrow i}  \right\rVert ^2_2 + \sum_{i=1}^M  \sum_{j=1, j \neq i}^M \left\lVert  \boldsymbol{W}^l_{i \rightarrow j}  \right\rVert ^2_2
\end{equation}

\noindent where $\alpha$ is a pre-defined parameter that determines the trade-offs between the penalties of intra-weights and inter-weights. In general, $\alpha$ is set to be smaller than 1, indicating that a smaller penalty is imposed on intra-weights, as compared with inter-weights. The reason is that the prediction of future demand for a service mode benefits more from the historical observations of its own features, than features of other service modes. 

Let $\mathcal{X}_m = \{\mathbf{X}_m^1, ..., \mathbf{X}_m^{N_m} \}$, $\mathcal{Y}_m = \{\mathbf{Y}_m^1, ..., \mathbf{Y}_m^{N_m} \} $ denote the training features and labels of task $m$ (the predicted demand for service mode $m$), where $N_m$ is the number of training samples of task $m$. In our problem, $N_m$ is the total number of time steps to be predicted in training dataset. Therefore, in a RCT learning framework, the parameters of the networks can be trained by solving the following problem: 

\begin{equation}  \label{eq:p1}
\min_{\mathcal{W}, \mathbf{b}} \sum_{m=1}^M \sum_{s=1}^{N_m} \left\lVert  \hat{ \mathbf{X}}_m^s  - \mathbf{X}_m^s \right\rVert ^2_2 + \beta_1 \sum_{l \in \mathcal{L} }J_1^l 
\end{equation}
where $\mathcal{L}$ is the set of layers, $\mathcal{W}, \mathbf{b}$ represent all weights and bias in parameters, $\hat{ \mathbf{X}}_m^s$ is the predicted value for ground truth $\mathbf{X}_m^s$ by the neural networks, $\beta_1$ is a parameter balancing the trade-offs between bias and variance. The first term minimizes the squared loss between predicted demand and actual demand, while the second term is a regularized term given by Eq. \ref{eq:j1}. 

\subsection{Multi-linear relationship learning}

\begin{figure}[!ht]
\centering
\includegraphics[width=0.62\textwidth]{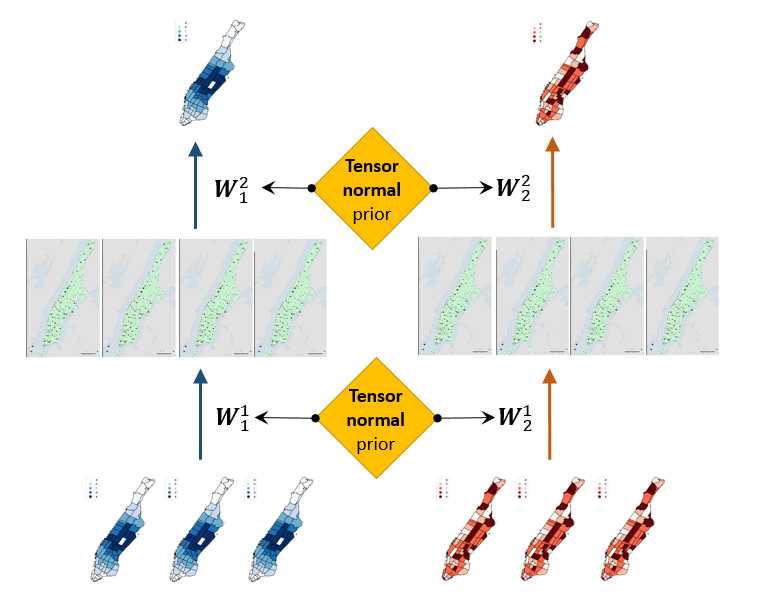}
\caption{Multi-linear learning}
\label{fig:multi-linear}
\end{figure}

In this section, we use an alternative weight to share knowledge across different tasks. As demonstrated in Fig. \ref{fig:multi-linear}, instead of building cross connections between the inputs and outputs of networks for different service modes, we apply a MLR learning module (first proposed by \cite{long2017learning}) that imposes a prior normal distribution on the intra-weights of multiple networks. This indicates that, the intra-weights of different networks are constrained by each other and subject to a common prior probability distribution.

First, we place the weights of all networks in layer $l$ in one tensor, denoted by $\mathcal{W}^l$, shown as follows: 
\begin{equation}
\mathcal{W}^l = [\boldsymbol{W}^l_{1 \rightarrow 1}, \boldsymbol{W}^l_{2 \rightarrow 2}, ...., \boldsymbol{W}^l_{M \rightarrow M}] \in \mathbb{R}^{\tilde{f_i}  \times f_o \times M}
\end{equation}

\noindent where $\tilde{f_i}, f_o$ are the input and output dimensions of one weight matrix as defined in Section IV.B, $M$ is the number of tasks (or service modes). Let $\mathcal{X} = \{ \mathcal{X}_m \}_{m=1}^{M}$, $\mathcal{Y} = \{ \mathcal{Y}_m \}_{m=1}^{M}$ denote the complete training data for all $M$ tasks. Given $\mathcal{X}$ and $\mathcal{Y}$, the Maximum A Posterior (MAP) estimation of parameters $\mathcal{W} = [..., \mathcal{W}^l, ...]$ is

\begin{equation} \label{eq:mae}
\begin{split}
p(\mathcal{W} | \mathcal{X}, \mathcal{Y}) & \propto p(\mathcal{W}) \cdot p(\mathcal{Y} | \mathcal{X}, \mathcal{W}) \\ 
&  = \prod_{l \in \mathcal{L}}  p(\mathcal{W}^l) \cdot \prod_{m =1}^{M} \prod_{n =1}^{N_m} p(\mathbf{Y}_m^n |\mathbf{X}_m^n, \mathcal{W}^l )
\end{split}
\end{equation}

\noindent where the first term in the right-hand-side, $p(\mathcal{W}^l) $, is the prior, and the second term,  $p(\mathbf{Y}_m^n |\mathbf{X}_m^n, \mathcal{W}^l )$, is a maximum likelihood estimation (MLE) given by the neural networks. We assume that the joint weight tensor $\boldsymbol{W}^l$ follows a tensor normal prior distribution as below,  

\begin{equation} \label{eq:prior}
\boldsymbol{W}^l \sim \mathcal{TN}_{\tilde{f_i} \times f_o \times M} (\overline{ \boldsymbol{W}}^l, \Sigma^l)
\end{equation}

\noindent where $\overline{ \boldsymbol{W}}^l$ is the mean tensor, $\Sigma^l \in \mathbb{R}^{(\tilde{f_i} \cdot f_o \cdot M) \times (\tilde{f_i} \cdot f_o \cdot M)}$ is the covariance matrix. As pointed out by \cite{long2017learning}, this assumption in the prior term can well capture the multi-linear relationship across parameter tensors. The covariance matrix $\Sigma^l $ may have an extreme large dimension, leading to computational difficulties. To address this issue, we decompose $\Sigma^l $ into the Kronecker product of three small covariance matrices: $\Sigma^l = \Sigma^l_I \otimes  \Sigma^l_O \otimes \Sigma^l_M $, where $\Sigma^l_I \in \mathbb{R}^{\tilde{f_i} \times \tilde{f_i}}$, $\Sigma^l_O  \in \mathbb{R}^{f_o \times f_o}$, $\Sigma^l_M \in \mathbb{R}^{M \times M}$ are input covariance matrix, output covariance matrix, and service mode covariance matrix, respectively. The input covariance matrix $\Sigma^l_I $ is computed by the covariance between the rows of the mode-1 matrix\footnote{
The $j$th row of mode-k matrix of the tensor $\boldsymbol{W}^l$, i.e. $\boldsymbol{W}^l_{(k)} $, contains all elements of  $\boldsymbol{W}^l$ with the $k$th index equal to $j$.} of $\boldsymbol{W}^l $, i.e. $\boldsymbol{W}^l_{(1)} \in \mathbb{R}^{\tilde{f_i} \times (f_o \cdot M)}$. The other two covariance matrices $\Sigma^l_O$ and  $\Sigma^l_M$ are computed in a similar way. 

Substituting Eq. \ref{eq:prior} into Eq. \ref{eq:mae} and taking the negative logarithm give rise to the following regularized optimization problem: 

\begin{equation} \label{eq:p2}
\min_{\mathcal{W}, \mathbf{b}} \sum_{m=1}^M \sum_{s=1}^{N_m} \left\lVert  \hat{ \mathbf{X}}_m^s  - \mathbf{X}_m^s \right\rVert ^2_2 + \frac{1}{2} \beta_2 \sum_{l \in \mathcal{L}} J_2^l 
\end{equation}
where $\beta_2$ is a parameter balancing the trade-offs between bias and variance, the regularized term $J_2^l $ in layer $l$ is given by, 
\begin{equation}
\begin{split}
J_2^l = & \text{vec}  (\boldsymbol{W}^l)^T (\Sigma^l_I \otimes  \Sigma^l_O \otimes \Sigma^l_M)^{-1}  \text{vec}  (\boldsymbol{W}^l) \\
&- \frac{D}{\tilde{f_i}} \text{ln} ( |\Sigma^l_I| ) - \frac{D}{f_o} \text{ln} ( |\Sigma^l_O| ) - \frac{D}{M} \text{ln} ( |\Sigma^l_M| ) 
\end{split}
\end{equation}

\noindent where $D=\tilde{f_i} \cdot f_o \cdot M$. The convariance matrices $\Sigma^l_I$,  $\Sigma^l_O$, $\Sigma^l_M$ are updated with the flip-flop algorithm \citep{ohlson2013multilinear}, during training process. In addition, we can fix $\Sigma^l_I$ and/or  $\Sigma^l_O$ (for example, assigned with identity matrices) and do not update their values during training process to increase training stability. In this condition, the model only focuses on the knowledge sharing across different tasks. Moreover, it can be found that the regularized terms in optimization problems \ref{eq:p1} and \ref{eq:p2} are layer separable. Therefore, we can design a multi-layer network that shares knowledge across tasks, with RCT learning in some layers and MLR learning in other layers. Mathematically, we can formulate a flexible network below, 

\begin{equation}
\min_{\mathcal{W}, \mathbf{b}} \sum_{m=1}^M \sum_{s=1}^{N_m} \left\lVert  \hat{ \mathbf{X}}_m^s  - \mathbf{X}_m^s \right\rVert ^2_2 
+ \beta_1 \sum_{l \in \mathcal{L}_c }J_1^l 
+\frac{1}{2} \beta_2 \sum_{l \in \mathcal{L}_m} J_2^l 
\end{equation}

\noindent where $\mathcal{L}_c$ and $\mathcal{L}_m$ are the set of layers using RCT and MLR learning, respectively.

\section{Experimental results}

\subsection{Data and models}
\begin{figure}[ht!]
\centering
\includegraphics[width=0.62\textwidth]{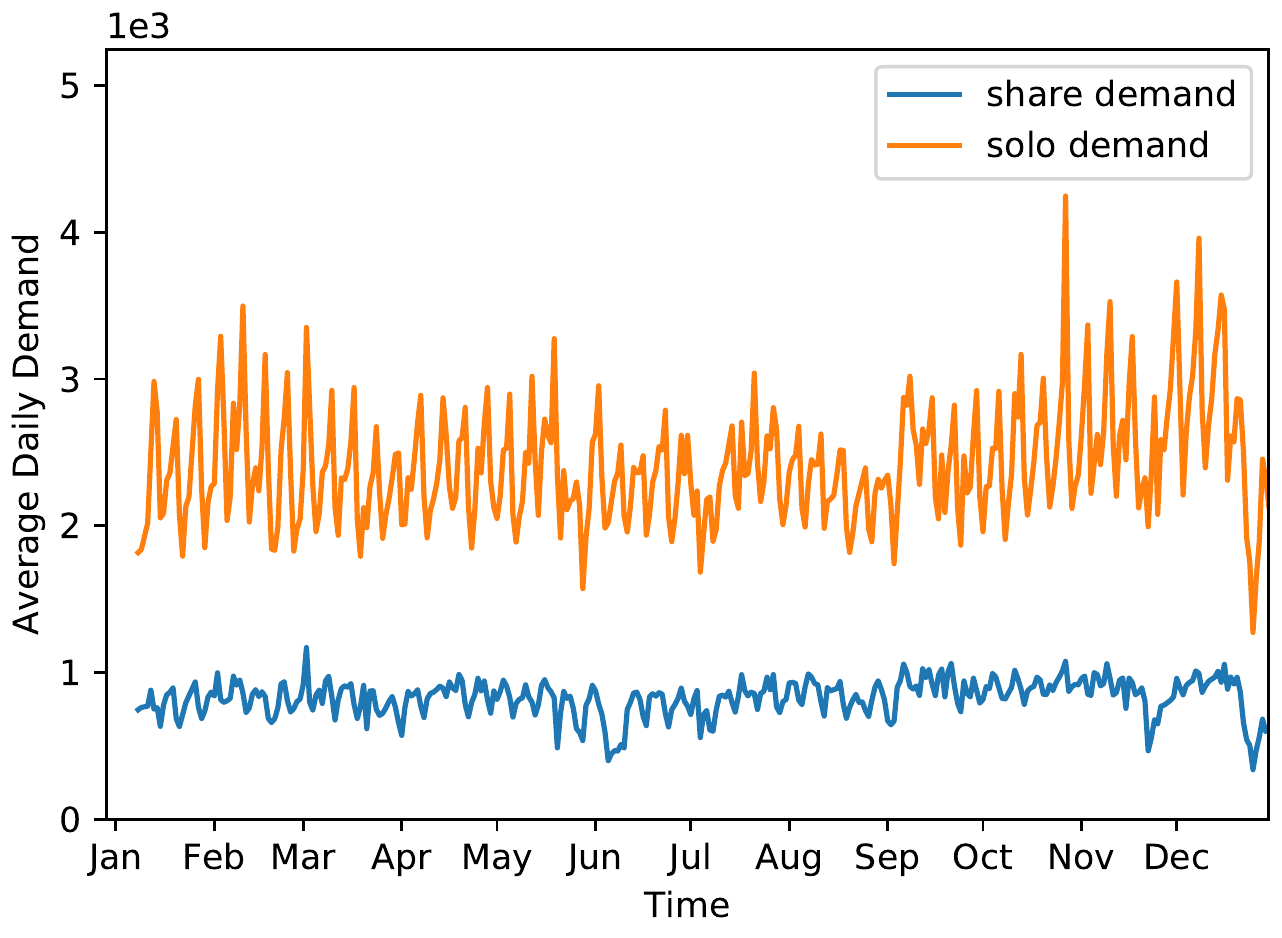}
\caption{Time series of Manhattan ride-hailing demand}
\label{fig:data-pattern}
\end{figure}

In September 2018, New York TLC released the new for-hire-vehicle data, which was reported by transportation network companies such as Uber and Lyft. The dataset includes detailed pick-up and drop-off time (on a basis of second) of the passengers as well as the TLC zone based pick-up and drop-off locations. In the dataset, there is a field representing the service mode of the trip, i.e., a solo ride or a shared ride. Based on this dataset, we summarize zone based hourly demand for both solo rides and shared rides in Manhattan (63 TLC zones in total). Fig. \ref{fig:data-pattern} illustrates the highly stochastic trend of daily demand for the two service modes in the year of 2018.

The spatial-temporal ride-hailing demand dataset is fused with land use attributes via another open source dataset -- Smart Location Database. The dataset provides zone based land use properties such as number of residences, number of employments, number of retail and point of interest, which are used to calculate pair-wise semantic relations between zones. 

\begin{figure}[ht!]
\centering
\begin{subfigure}[b]{0.36\textwidth}
\includegraphics[width=\textwidth]{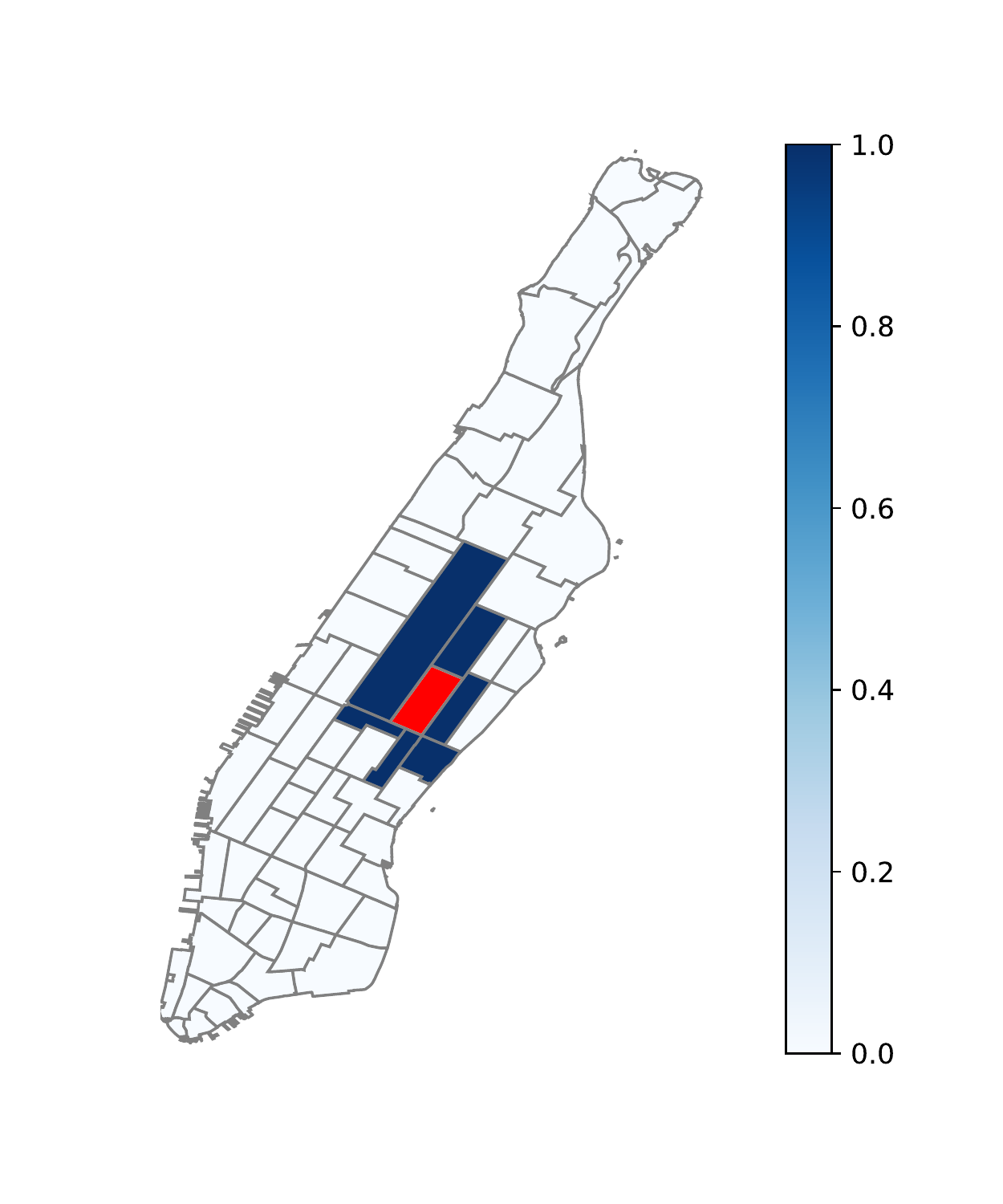}
\caption{neighbor graph}
\label{fig:graph-nei}
\end{subfigure}
\begin{subfigure}[b]{0.36\textwidth}
\includegraphics[width=\textwidth]{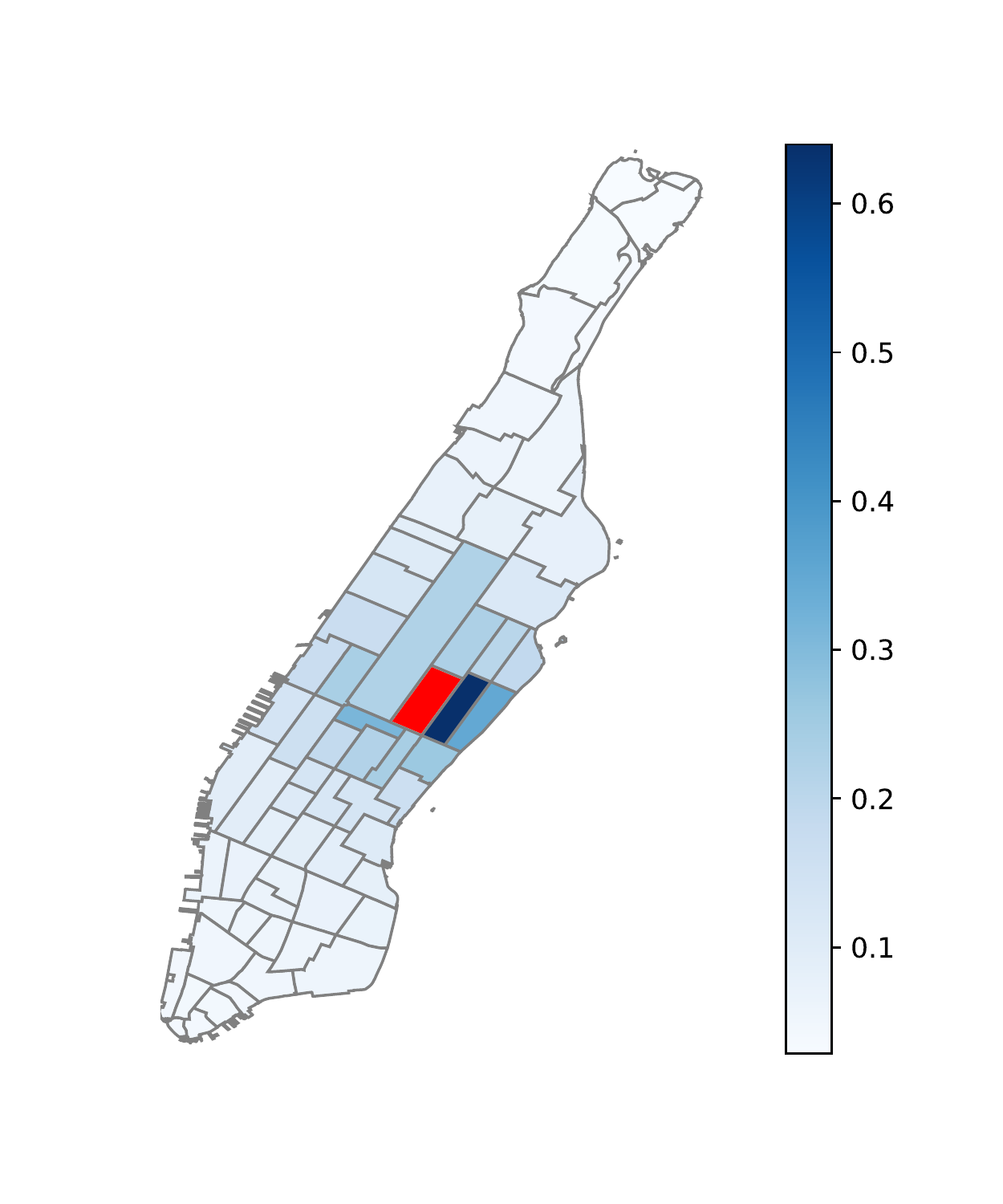}
\caption{distance graph}
\label{fig:graph-dis}
\end{subfigure}
\\
\begin{subfigure}[b]{0.36\textwidth}
\includegraphics[width=\textwidth]{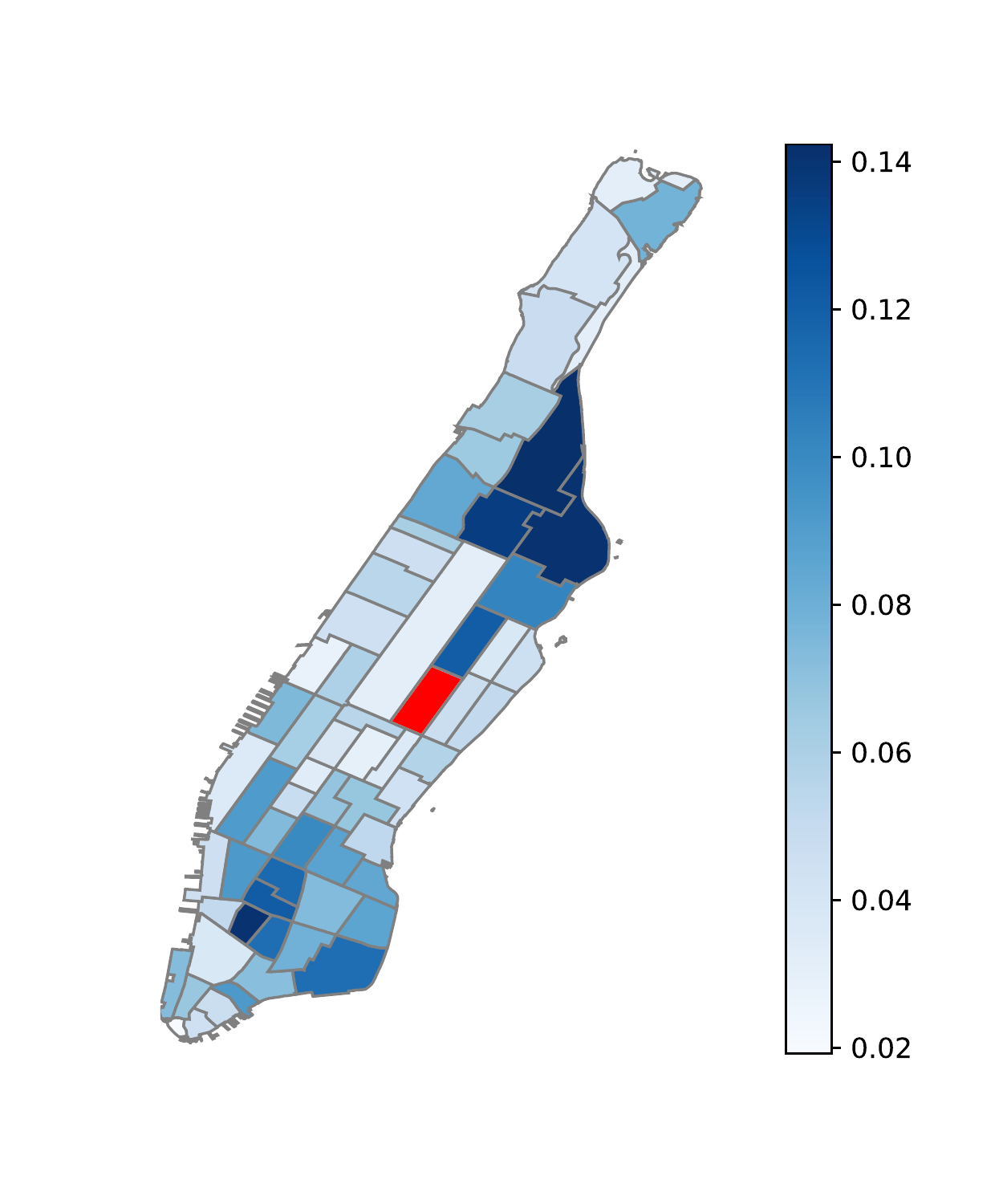}
\caption{functionality graph}
\label{fig:graph-fun}
\end{subfigure}
\begin{subfigure}[b]{0.36\textwidth}
\includegraphics[width=\textwidth]{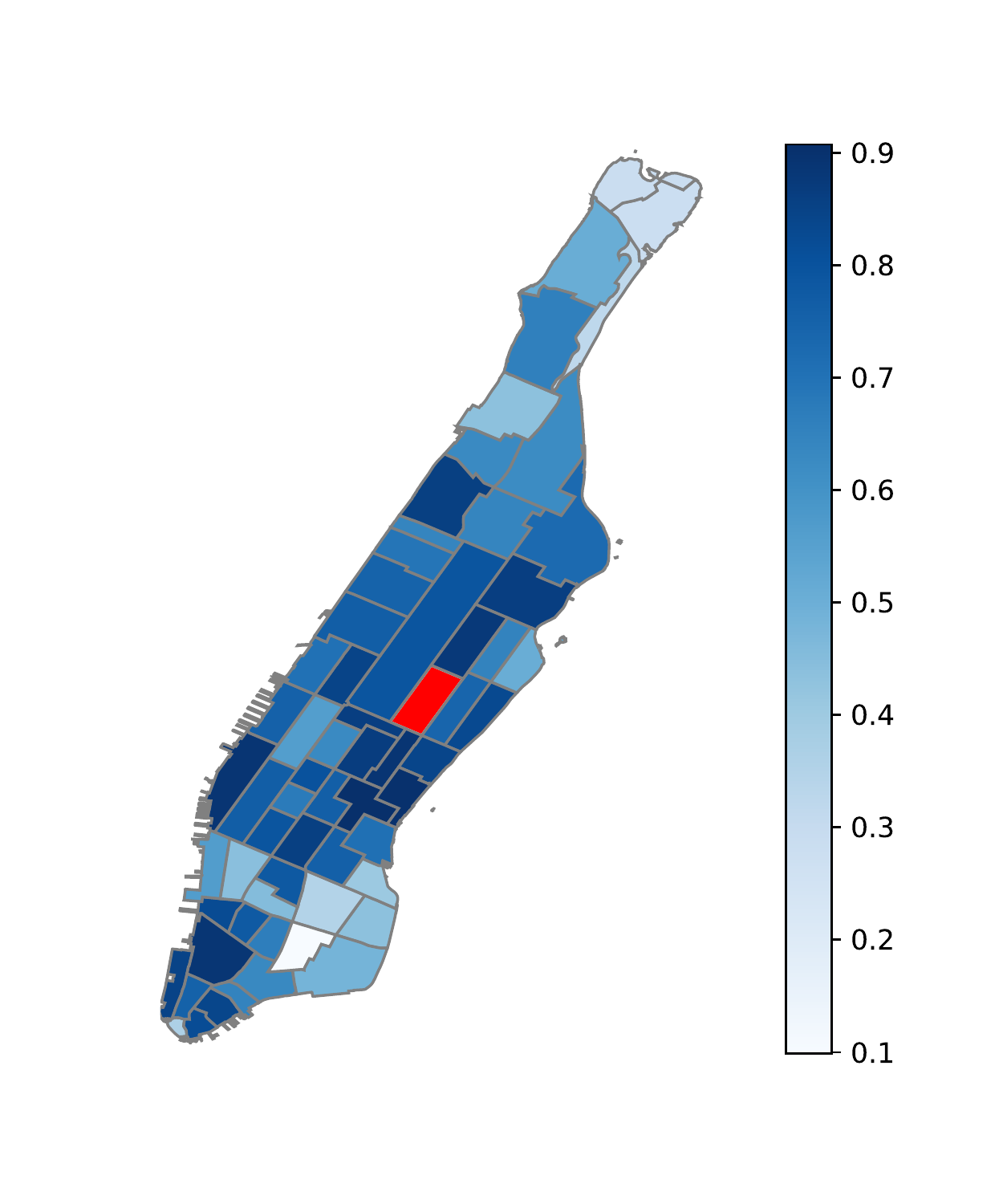}
\caption{shared graph}
\label{fig:graph-sha}
\end{subfigure}
\caption{Graphs of zone 237}
\label{fig:graph}
\end{figure}

With the aforementioned spatial dependence (i.e. graphs), Fig. \ref{fig:graph} presents the multi-graph of zone 237 as an example. The target zone (id 237) is marked with red color. All the adjacent zones are highlighted in Fig. \ref{fig:graph-nei}. The distance graph is shown in Fig. \ref{fig:graph-dis}, in which zones closer to 237 have a higher value. The functionality graph calculated in Eq.\ref{eq:fun} is illustrated in Fig.\ref{fig:graph-fun}, and the spatial correlation of shared service demand is shown in Fig.\ref{fig:graph-sha}. Neighbor and distance can only capture the spatial dependence of nearby zones; unlikely, some distant zones may have strong correlation in functionality or shared service demand pattern. The non-Euclidean correlation can provide useful information in our spatial-temporal prediction problem.

In this real-world experiment, we use the demand data from 8 January 2018 to 4 November 2018 for models' training, 5 November 2018 to 2 December 2018 for models' validation, and 3 December 2018 to 31 December for models' testing. We compare different state-of-art machine learning approaches with the proposed deep multi-task learning approaches in terms of prediction accuracy. The models considered in this paper are described below: 

\begin{itemize}
\item \textbf{RCT-MGC}: a deep learning model that uses two symmetric four-layer MGC networks (with 128, 256, 128, 1 units) for the two prediction tasks (solo and shared service demand). The four layers in the two networks are connected with RCT modules. 
\item \textbf{MLR-MGC}: a deep learning model that builds similar structure with RCT-MGC, except that the four layers in the networks for the two tasks share knowledge with each other with MLR.  
\item \textbf{MIX-MGC}: a deep learning model that has similar structure with RCT-MGC, except that the two lower layers share knowledge through RCT and the two upper layers share knowledge through MLR. 
\item \textbf{MGC}: MGC networks without multi-task learning. 
\item \textbf{MLP}: multi-layer perception model, which is a basic neural network containing at least three layers. 
\item \textbf{XGB}: XGBoost model, which is an implementation of gradient boosted decision trees. 
\item \textbf{GBDT}: gradient boosting decision tree model, which constructs multiple regression decision trees. 
\item \textbf{RF}: random forest model trained by bootstrapped samples of each decision tree. 
\item \textbf{LASSO}: LASSO model with historical demand as input and absolute shrinkage and selection operator in the loss function. 
\item \textbf{HA}: the average historical demand over the past four weeks. 
\end{itemize}

The parameters of all abovementioned state-of-art machine leanring models are fine-tuned. Each model is fed with features including $\boldsymbol{X}_{m}^{t-1}$, $\boldsymbol{X}_{m}^{t}$ (the most recent two historical demands), $\boldsymbol{X}_{m}^{t+1-24}$ (historical demands during the same hour on yesterday), and $\boldsymbol{X}_{m}^{t+1-24 \times 7}$ (historical demands during same hour on last week). In RCT-MGC, the hyperparameter $\alpha$ is 0.1 to impose relatively small penalty on the intra-weights, and relatively large penalty on the inter-weights. The two balancing factors in the objective function $\beta_1$ and $\beta_2$ are 0.001 and 0.1 respectively. The neural networks are implemented using pytorch and optimized via Adam optimizer with a learning rate of 0.001 and a batch size of 16. All experiments are implemented on a server with 64G RAM and one NVIDIA 1080Ti GPU.

\subsection{Results on the testing dataset}

We examine the prediction error of the models by three measurements, Root Mean Square Error (RMSE), Mean Absolute Error (MAE) and Mean Absolute Percentage Error (MAPE). Since a zero observed hourly demand will drive MAPE to infinity, we only include the data records with positive demand for the calculation of MAPE. The performances of the models are depicted in TABLE \ref{table:test-error}. For both solo service demand and shared service demand, the four deep learning models significantly outperform the benchmarks of conventional machine learning models. For instance, compared with the MLP model, the MGC model can reduce RMSE/MAE/MAPE by 12.4\%/14.1\%/11.7\% for solo service demand prediction, and reduce the measurements by 10.0\%/10.0\%/17.5\% for shared service demand prediction. This indicates that the spatial correlations (i.e., both Euclidean and non-Euclidean dependencies) provide important information in spatial-temporal ride-hailing demand prediction; the correlations can be well characterized by the proposed adjacent matrices in the MGC modeling framework. 

Moreover, based on the comparison between model MGC and models RCT-MGC, MLR-MGC and MIX-MGC, we note that a multi-task learning structure can further improve the prediction accuracy. The results indicate that demands of different ride-hailing service modes have notable correlations, which can be captured via deep multi-task learning approaches. Additionally, since both MLR-MGC and MIX-MGC are better than RCT-MGC for both solo service demand and shared service demand, we may conclude that a MLR structure fits the dataset better than a RCT structure. 

\begin{table}[ht]
    \footnotesize
    \centering
    \caption{Results of the testing dataset }
    \begin{tabular}{p{1.8cm}p{1.1cm}p{1.1cm}p{1.1cm}} \hline
    \multicolumn{4}{c}{demand of solo service rides} \\ [0.5ex] \hline
    Model & RMSE & MAE & MAPE  \\ [0.5ex] \hline
    RCT-MGC & 20.238 & 12.949 & 0.216 \\ 
    MLR-MGC & 19.896 & 12.963 & 0.239 \\
    MIX-MGC & 19.726 & 12.748 & 0.235 \\
    MGC & 20.555 & 13.097 & 0.226 \\ 
    MLP & 23.459 & 15.246 & 0.256 \\ 
    XGB & 23.721 & 15.334 & 0.256 \\ 
    GBDT & 23.806 & 15.365 & 0.256 \\ 
    RF & 24.623 & 15.908 & 0.260 \\ 
    LASSO & 26.906 & 17.365 & 0.308 \\
    HA & 53.712 & 29.835 & 0.471 \\ \hline
    \multicolumn{4}{c}{demand of shared service rides} \\ [0.5ex] \hline
    Model & RMSE & MAE & MAPE \\ [0.5ex] \hline
    RCT-MGC & 9.316 & 6.059 & 0.322 \\ 
    MLR-MGC & 8.937 & 5.994 & 0.343 \\
    MIX-MGC & 8.727 & 5.919 & 0.343 \\
    MGC & 9.536 & 6.346 & 0.350 \\ 
    MLP & 10.595 & 7.050 & 0.424 \\ 
    XGB & 10.621 & 6.999 & 0.401 \\ 
    GBDT & 10.670 & 7.017 & 0.401 \\ 
    RF & 11.187 & 7.405 & 0.420 \\ 
    LASSO & 12.465 & 7.986 & 0.476 \\
    HA & 19.227 & 10.931 & 0.600 \\ \hline
    \end{tabular}
    \label{table:test-error}
\end{table}

\begin{figure}[ht!]
\centering
\begin{subfigure}[b]{0.58\textwidth}
\includegraphics[width=\textwidth]{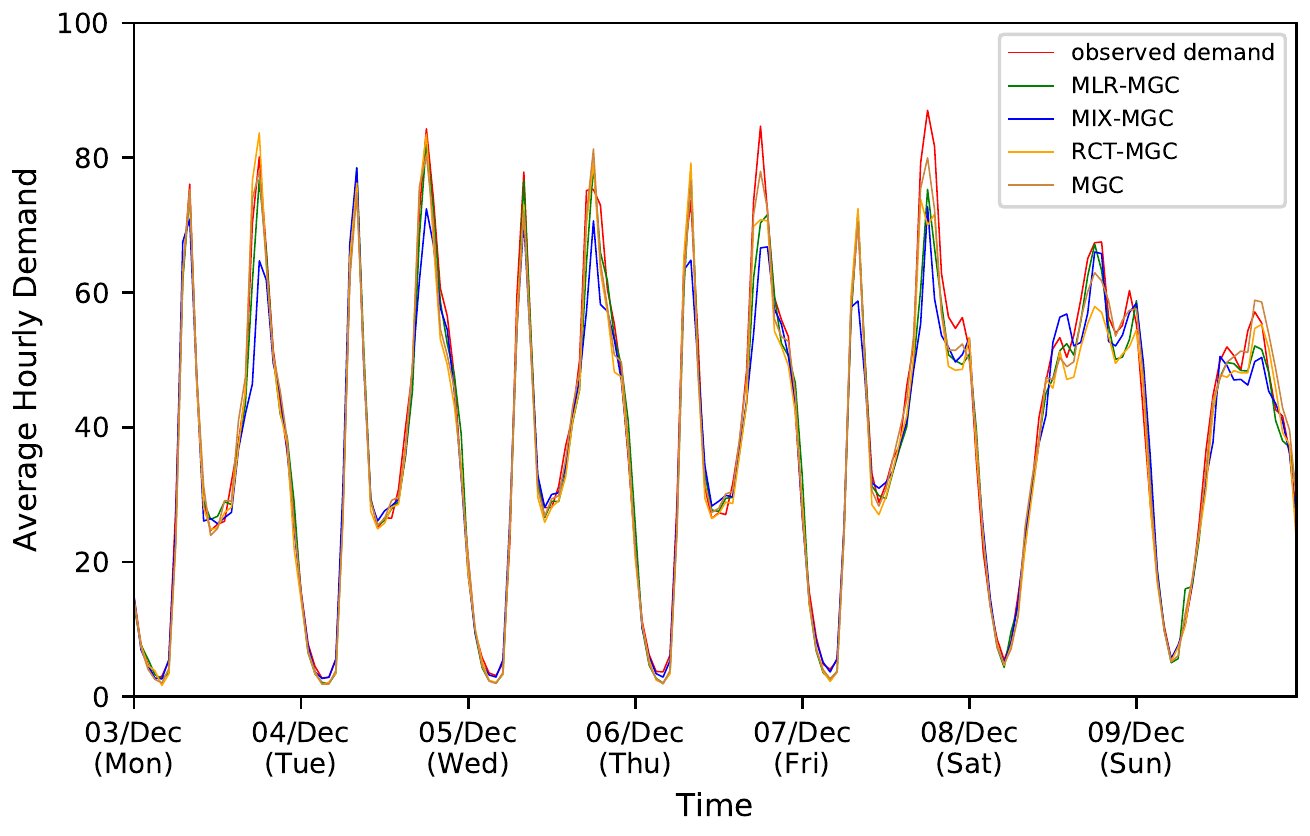}
\caption{Regular day}
\label{fig:pred-reg}
\end{subfigure}
\\
\begin{subfigure}[b]{0.58\textwidth}
\includegraphics[width=\textwidth]{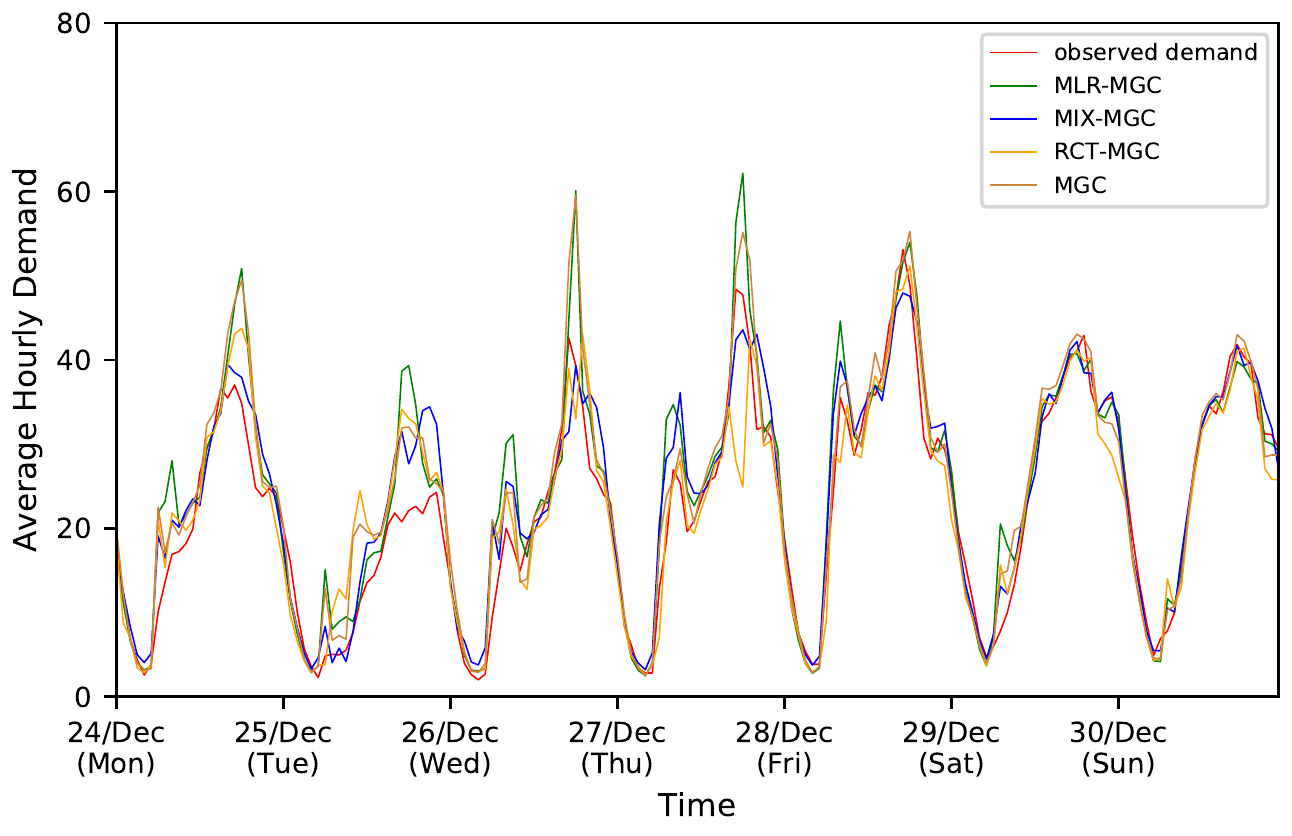}
\caption{Holiday day}
\label{fig:pred-holi}
\end{subfigure}
\caption{Hourly prediction results}
\label{fig:pred}
\end{figure}

Fig. \ref{fig:pred} depicts  hourly prediction results of shared service demand versa real-world observations. In both regular days (Fig. \ref{fig:pred-reg}) and holidays (Fig. \ref{fig:pred-holi}), the deep learning models can accurately capture the upcoming demand.

\section{Conclusion}
This paper studies the joint prediction of passenger demands for multiple service modes in ride-hailing systems. To enable effective knowledge sharing across different spatial-temporal prediction tasks, We propose a novel deep multi-task multi-graph learning approach, which first establishes separate MGC networks for different service modes, and then connects the networks with RCT and MLR learning techniques. While RCT learning builds up concrete bridges between different MGC networks, MLR learning imposes a soft connection among various MGC networks by assuming that their parameters follow a common prior probability distribution. Evaluated against a real-world ride-hailing dataset in Manhattan, we show that our proposed models significantly outperform the benchmark algorithms. Moreover, the use of multi-task learning techniques on the basis of MGC networks can further improve the prediction accuracy in spatial-temporal prediction tasks for multiple service modes. This study opens a few avenues that worth exploration, to name a few, (1) joint predictions of passenger demands for different transportation modes (such as bikes, private cars, and public transits); (2) joint predictions of passenger demand for ride-hailing sevices on multi-zone levels.   

\section*{Acknowledgment}
The work described in this paper was supported by a grant from Hong Kong Research Grants Council under project HKUST16208619 and a NSFC/RGC Joint Research grant N$\_$HKUST627/18. This work was also supported by the Hong Kong University of Science and Technology - DiDi Chuxing (HKUST-DiDi) Joint Laboratory.

\bibliographystyle{apalike}
\bibliography{references}{}




\end{document}